\documentclass[lettersize,journal]{IEEEtran}
\usepackage{amsmath,amsfonts}
\usepackage{algorithmic}
\usepackage{algorithm}
\usepackage{array}

\usepackage{textcomp}
\usepackage{stfloats}
\usepackage{url}
\usepackage{verbatim}
\usepackage{graphicx}
\usepackage{cite}
\usepackage{booktabs}
\usepackage{bm}
\usepackage{multirow}

\usepackage{tikz}
\usepackage{comment}
\usepackage{amsmath,amssymb} 
\usepackage{color}
\definecolor{dodgeblue}{RGB}{30,144,255}
\definecolor{lightgreen}{RGB}{0,157,0}

\usepackage{diagbox}

\usepackage[breaklinks=true,colorlinks,bookmarks=false]{hyperref}

\usepackage{colortbl}
\definecolor{mygray}{gray}{.9}
\usepackage{array, boldline, makecell, booktabs}

\usepackage{xcolor}
\definecolor{firered}{RGB}{222,82,57}
\definecolor{iceblue}{RGB}{33,102,200}
\usepackage{multirow}
\usepackage{makecell}
\usepackage{diagbox} 
\definecolor{mygray}{gray}{.9}
\usepackage{adjustbox}
\usepackage{graphicx}
\usepackage{bm}
\usepackage{subfigure}
\usepackage{algorithmic}
\usepackage{multirow}
\usepackage{colortbl}
\usepackage{booktabs}
\usepackage{array}
\usepackage{color}
\definecolor{tabcolor}{rgb}{0.84,0.87,0.96}

\usepackage{textcomp}
\usepackage{stfloats}
\usepackage{bm}
\makeatletter
\newcommand{\thickhline}{%
    \noalign {\ifnum 0=`}\fi \hrule height 1pt
    \futurelet \reserved@a \@xhline
}

\begin{document}

\title{SSGA-Net: Stepwise Spatial Global-local Aggregation Networks for for Autonomous Driving}

\author{Yiming Cui, Cheng Han, Dongfang Liu 

\IEEEcompsocitemizethanks{
\IEEEcompsocthanksitem Yiming Cui is with the University of Florida, Gainesville, FL 32611, USA.
\IEEEcompsocthanksitem Cheng Han and Dongfang Liu are with the Rochester Institute of Technology, Rochester, New York, USA.}
}
\maketitle
\vspace{-2.6cm}
\begin{abstract}
 Visual-based perception is the key module for autonomous driving. Among those visual perception tasks, video object detection is a primary yet challenging one because of feature degradation caused by fast motion or multiple poses. Current models usually aggregate features from the neighboring frames to enhance the object representations for the task heads to generate more accurate predictions. Though getting better performance, these methods rely on the information from the future frames and suffer from high computational complexity. Meanwhile, the aggregation process is not reconfigurable during the inference time. These issues make most of the existing models infeasible for online applications. To solve these problems, we introduce a stepwise spatial global-local aggregation network. Our proposed models mainly contain three parts: 1). Multi-stage stepwise network gradually refines the predictions and object representations from the previous stage; 2). Spatial global-local aggregation fuses the local information from the neighboring frames and global semantics from the current frame to eliminate the feature degradation; 3). Dynamic aggregation strategy stops the aggregation process early based on the refinement results to remove redundancy and improve efficiency. Extensive experiments on the ImageNet VID benchmark validate the effectiveness and efficiency of our proposed models.
\end{abstract}
%
%
%
\vspace{-0.8cm}

\section{Introduction}
With the prices of storage devices decreasing, more videos have been captured and preserved in recent years. Regarding autonomous driving applications, single images may not provide enough information to make decisions and predictions. Therefore, instead of only focusing on image analysis, researchers have begun investigating how to recognize objects in the video frames \cite{yang2018efficient,  kim2020video, cui2022dg, voigtlaender2019mots, yang2019video}.  For example, multiple tasks together with their corresponding datasets are proposed in the field of video analysis, like video object detection \cite{russakovsky2015imagenet, zhang2018improved, zhu2017deep, cui2023learning}, video object segmentation \cite{yang2018efficient, caelles2017one}, video instance segmentation \cite{yang2019video,liu2021sg, Cao_SipMask_ECCV_2020} and multi-object tracking and segmentation \cite{cui2022dg, voigtlaender2019mots, yang2020remots, cui2024collaborative}. Among them, video object detection is the fundamental yet challenging task \cite{zhu2017deep, zhu2018towards}, which needs to handle the situations that rarely occur in the image domains like motion blur, defocus, and so on. To handle the feature degradation issues mentioned above, one commonly used method is post-processing \cite{han2016seq, kang2016object, kang2017t, sabater2020robust, song2022cross}. These models first apply the image object detectors to each frame and then use the hints provided by the video temporal information, like optical flow, to associate the predictions. However, these models are not trained end-to-end, which makes their performance lower than the other models like feature-aggregation-based methods.

Besides post-processing-based models, another direction used for video object detection is aggregating the features for video object detection \cite{chen2020memory, zhu2017flow, wu2019sequence, gong2021temporal, wang2023sac, leng2023error, wang2018fully, deng2019relation, han2020mining, cui2021tf, lin2020dual}. In detail, these methods usually fuse the features of the neighboring frames according to their semantic similarity \cite{wu2019sequence, jiang2020learning, han2020mining, deng2019relation, chen2020memory} or temporal relations \cite{wang2018fully, zhu2017flow, zhu2018towards}. Regarding the way of fusion, these models can also be divided into two groups: weighted averaging \cite{zhu2017flow, wang2018fully, chen2020memory, wu2019sequence}, or learnable networks \cite{cui2021tf, he2021end, zhou2022transvod, wang2022ptseformer, cui2023faq}. Built based on classic image object detectors, these feature-aggregation-based models usually pay more attention to the temporal global or local or both aggregation rather than the spatial domain. With Transformer-based object detectors \cite{detr,zhu2021deformable, liu2022dab, gao2021fast, li2022dn, chen2022group, zhang2022dino} introduced to the video domain, recent works have begun to fuse both spatial and temporal features globally and locally, like TransVOD \cite{zhou2022transvod, he2021end} and PTSEFormer \cite{wang2022ptseformer}. However, the performance of these methods is highly influenced by the number of frames used for aggregation \cite{zhu2017flow, zhu2018towards, cui2022dfa}. Moreover, these Transformer-based video object detectors are not reconfigurable and always need to be retrained when the number of frames used for aggregation is modified.


Though getting a good performance on the video object detection task, those methods mentioned above are not suitable for online video applications because they either have a high computational complexity \cite{he2021end, zhou2022transvod,wang2022ptseformer,gong2021temporal} or require future temporal information \cite{chen2020memory, jiang2020learning, deng2019relation, han2020mining} during the inference process. Therefore, recent works have begun to focus on improving the current models' efficiency and applying them to online applications. For a better inference speed, some work \cite{chen2021exploring, sabet2021temporal, Xu2020CenterNetHP, sabater2020robust, sun2022efficient} replace the two-stage object detectors with a one-stage version. Other works \cite{cui2022dynamic, jiang2020learning} improve the model efficiency by dynamically sampling the frames used for feature aggregation. 

\begin{figure*}[!bt]
    \centering
    \subfigure[]{
         \includegraphics[width=5cm]{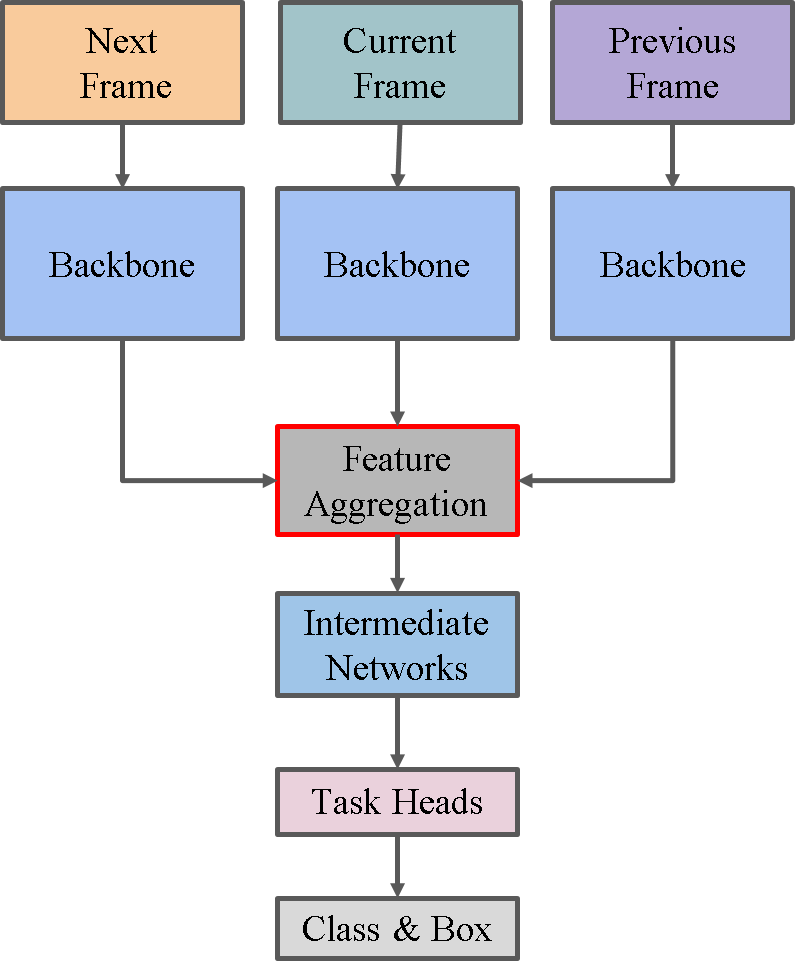}
     }
     \subfigure[]{
         \includegraphics[width=9cm]{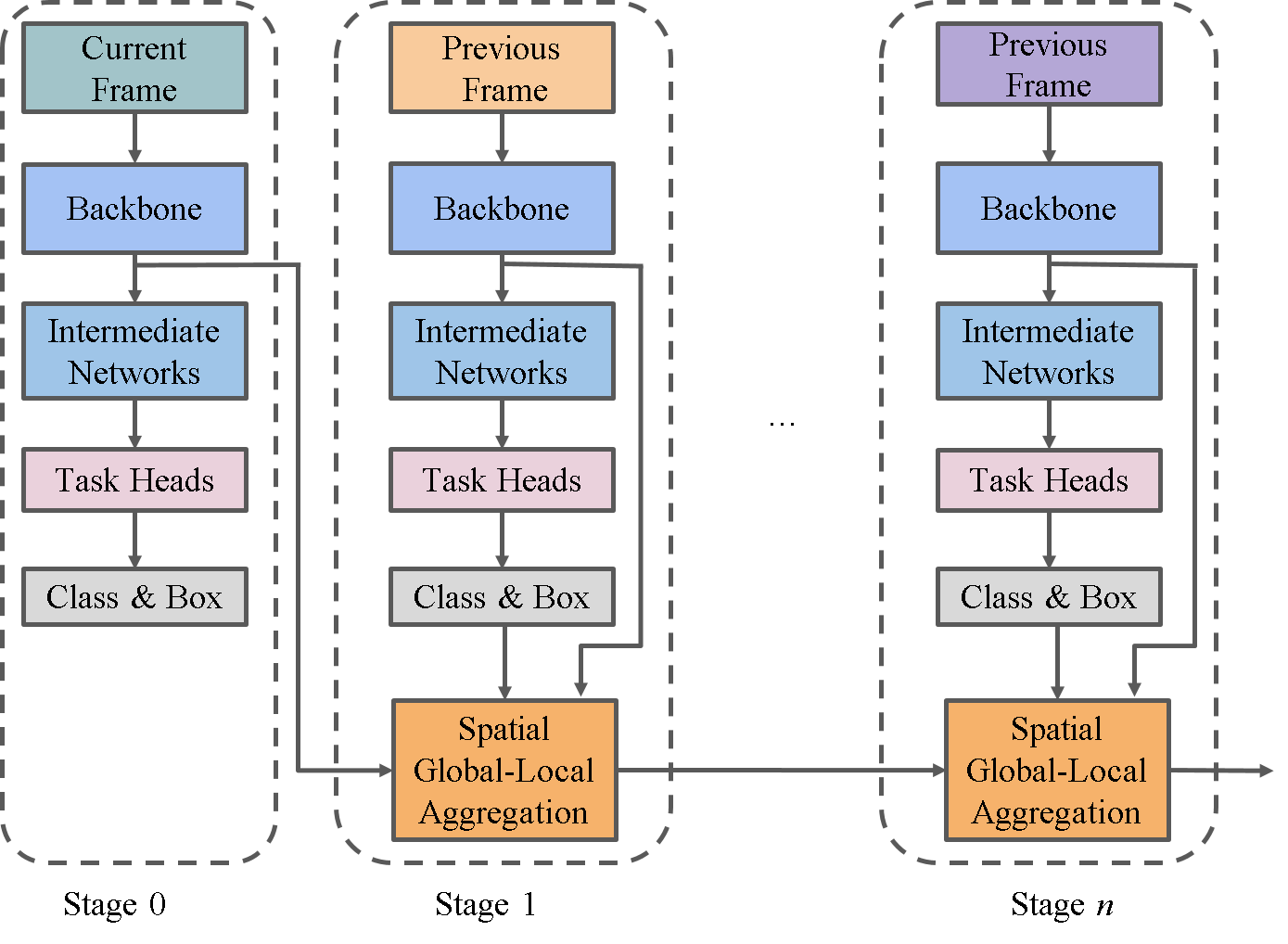}
         }
    \caption{Comparison of model frameworks between the existing feature-aggregation based models (a), which fuse the features at one time, and our proposed models (b), which gradually refine the results. }
    \label{fig: modelCompare}
\end{figure*}

To summarize, most of the existing post-processing-based video object detectors can be updated for online applications but suffer from low performance. When it comes to the feature-aggregation-based models, their high computational complexity limits their uses for online video object detection. Though the recent Transformer-based models perform better, their lack of reconfigurability, high computational complexity, and requirement for future frames make them unfeasible for online video applications. Figure \ref{fig: modelCompare} shows an example of a comparison between the existing feature-aggregation-based models and the approach we will propose. Different from the existing models in Figure \ref{fig: modelCompare}(a), which fuse all the features from the neighboring frames at one time, our approach, as shown in Figure \ref{fig: modelCompare}(b), refines the object representations and predictions gradually and is more suitable for the online video applications. Our contributions are summarized as follows:

\begin{itemize}
    \item We introduce a stepwise aggregation network that refines the prediction results gradually based on the neighboring frames. Moreover, a dynamic aggregation strategy is introduced to remove redundant aggregation to balance efficiency and performance.
    \item A spatial global-local aggregation module is introduced to enhance the object representations based on the enhanced global features from the previous stage and local features from the neighboring frame. This module only relies on the predictions and feature representations from the previous frames, which can be implemented with a queue for better efficiency and reconfigurability. Therefore, our proposed models are suitable for online video object detection. 
    \item Our module is a plug-and-play module and can be integrated into most of the existing Transformer-based object detectors for the online video object detection task. Evaluated on the ImageNet VID  benchmark, the performance can be improved by at least $1.0\%$ on mAP with our module integrated.
\end{itemize}

\section{Related Works}
\noindent\textbf{Visual perception for autonomous driving.} Visual perception is a crucial component of autonomous driving, and many research efforts \cite{9411961, liu2020indoor, dong2023watchdog, liu2020video, wang2022towards, ZHANG2018420, liu2020large} have been made to improve the performance of perception systems. 
In this section, we briefly review the related works about autonomous driving.

\noindent\textbf{Image object detection.} Object detection has traditionally relied on two-stage methods like R-CNN \cite{ren2015faster, he2017mask, girshick2015fast} that use predefined anchor boxes to identify candidate objects. However, to improve efficiency and speed, one-stage methods have been introduced that do not require region proposals, like FCOS \cite{tian2020fcos}, YOLO series \cite{redmon2018yolov3, wang2024yolov10, redmon2017yolo9000} and point/center-based models \cite{duan2019centernet, law2018cornernet}. More recently, query-based methods such as DETR \cite{detr}, Deformable-DETR \cite{zhu2021deformable}, and Sparse R-CNN \cite{sun2021sparse} have emerged. These methods \cite{detr, gao2022adamixer, roh2021sparse, gao2021fast, liu2022dab, zhang2022dino, Meng_2021_ICCV} replace anchor boxes and region proposals with learned proposals/queries and use transformers to generate a sequence of prediction outputs. As a pioneer work, DETR \cite{detr} uses a set loss function to perform bipartite matching between predicted and ground-truth objects, while Deformable-DETR \cite{zhu2021deformable} improves convergence by enhancing feature spatial resolutions. Following DETR and Deformable-DETR, multiple DETR-series models \cite{gao2021fast, roh2021sparse, Meng_2021_ICCV, zhang2022dino, liu2022dab} have been introduced to improve the efficiency and accuracy of Transformer-based object detectors. Different from the DETR-series models which are based on Transformers, Sparse R-CNN \cite{sun2021sparse} introduces a fixed sparse set of learned object proposals to classify and localize objects with dynamic heads, generating final predictions directly without non-maximum suppression post-processing methods \cite{bodla2017soft, liu2019adaptive}. These recent developments in object detection show promising results in improving efficiency, speed, and accuracy.

\noindent\textbf{Image instance segmentation.} Instance segmentation, an extension of object detection, requires algorithms to assign every object with a pixel-level mask. To achieve this, methods such as Mask R-CNN \cite{he2017mask} and YOLACT \cite{yolact-iccv2019} have been introduced, which add a mask head to existing object detection frameworks to produce instance-level masks. One-stage instance segmentation frameworks such as SipMask \cite{Cao_SipMask_ECCV_2020} and SOLO \cite{wang2020solo, wang2020solov2} have also been developed to balance inference time and accuracy. Recently, QueryInst \cite{Fang_2021_ICCV} has extended the query-based object detection method Sparse R-CNN  \cite{sun2021sparse} to instance segmentation by adding a dynamic mask head and parallel supervision. However, these two-stage and query-based methods have a fixed number of proposals, which can be problematic for images with varying numbers of objects and devices with different computational resource constraints. To address this issue, researchers are exploring more adaptive methods for instance segmentation. For example, BlendMask \cite{chen2020blendmask} uses a dynamic mask head and feature fusion mechanism to produce high-quality masks with fewer parameters.

\noindent\textbf{Video object detection.} Video object detection aims to categorize and localize objects of interest from each frame regardless of the potential feature degradation caused by the fast motion. Current methods \cite{kang2016object, han2016seq, kang2017t, wu2019sequence, zhu2017flow, zhu2018towards, chen2020memory, wang2018fully, cui2021tf} usually extend the image object detectors to the video domains. Depending on the way of extension, these models can be divided into two groups: Post-processing based and feature-aggregation based.

Post-processing-based models aim to extend image object detectors to the video domain by associating the prediction results in each frame according to temporal relations \cite{kang2016object, han2016seq, kang2017t}. These models, such as T-CNN \cite{kang2016object}  and Seq-NMS \cite{han2016seq}, have been developed to improve video object detection by using simple object tracking or associating prediction results based on IOU threshold. In detail, T-CNN \cite{kang2016object} introduces a CNN-based pipeline with simple object tracking for video object detection. Seq-NMS \cite{han2016seq} associates the prediction results from each frame according to the IOU threshold. Though getting a better performance than single-image object detectors, these models highly rely on the detection results on individual frames, which cannot be optimized jointly. Once the outputs of the single frame are wrong, the post-processing pipelines cannot update the predictions. However, these models are highly dependent on the detection results of individual frames, and they cannot be optimized jointly. If the outputs of the single frame are wrong, the post-processing pipelines cannot update the predictions, leading to suboptimal performance. Moreover, these models are typically slower since they rely on processing each frame separately before post-processing. 

In contrast, feature-aggregation-based and Transformer-based models can learn to aggregate information across frames and optimize the predictions jointly, leading to improved performance. These models are capable of utilizing temporal and spatial information to track and detect objects across consecutive frames, making them more suitable for video object detection tasks. Overall, while post-processing-based models can improve performance compared to single-image object detectors, they are limited by their dependence on individual frame detection results and slower inference speed. Feature-aggregation-based and Transformer-based models offer a more promising solution to video object detection tasks, leveraging information from multiple frames and optimizing jointly.

Feature-aggregation-based models aim to enhance the representations of the current frame by using the features of neighboring frames. This approach is based on the assumption that neighboring frames can help improve the current frame's features in case of feature degradation. Several models have been proposed to implement this idea. For instance, FGFA \cite{zhu2017flow} uses estimated optical flow to fuse neighboring features. MANet aggregates the features of objects on both pixel-level and instance-level. In contrast, SELSA \cite{wu2019sequence} calibrates the features based on semantic similarity instead of temporal relations. MEGA \cite{chen2020memory} combines the previous methods by aggregating local and global temporal information to boost performance. These models have shown improved performance compared to post-processing-based models. However, feature-aggregation-based approaches generally require more computational resources, leading to slower inference speeds. Overall, feature-aggregation-based models have been successful in improving video object detection performance. However, they usually come at the cost of slower inference speeds. Transformer-based models offer a promising solution to this issue by providing an efficient way to fuse information across frames. The development of these models is expected to lead to further improvements in video object detection tasks.

The current state-of-the-art methods for object detection, such as Faster R-CNN \cite{ren2015faster} and CenterNet \cite{duan2019centernet}, are mainly designed for image-based tasks. However, with the emergence of the DETR series, including DETR \cite{detr}, Fast-DETR\cite{gao2021fast}, Deformable-DETR \cite{zhu2021deformable}, and Sparse-DETR \cite{roh2021sparse}, researchers have started to explore the use of Transformer-based models for video object detection. One promising line of research is the TransVOD series \cite{zhou2022transvod, he2021end}, which introduces two key modules to enhance detection performance. The first is the Temporal Query Encoder, which fuses object queries across frames to capture temporal relationships. The second is the Temporal Deformable Transformer Decoder (TDTD), which generates detection results for the current frame. Similarly, the PTSEFormer \cite{wang2022ptseformer} proposes the Spatial Transition Awareness Model to fuse spatial and temporal information and improve predictions. These models still follow the feature-aggregation-based prototype, where features from neighboring frames are aggregated to enhance the features of the current frame before being fed into the task heads for the final prediction. Overall, these approaches show great potential for advancing video object detection and may lead to even more accurate and efficient models in the future.

\noindent\textbf{Video instance segmentation.} Video instance segmentation is a challenging task that requires the association of segmented instances across multiple frames. To accomplish this, it is crucial to fully utilize the temporal information in the video sequence and establish the spatiotemporal relationship between pixels across frames. Currently, there are two paradigms for building video instance segmentation (VIS) models: multiple-stage and single-stage.

Multiple-stage approaches divide the problem into two components: detection and association. The baseline method, MaskTrack R-CNN \cite{yang2019video}, extends the two-stage image instance segmentation model, Mask R-CNN \cite{he2017mask}, with a tracking branch to implement single-frame instance segmentation and inter-frame tracking. Most previous studies have followed the tracking-by-segmentation pipeline to associate instances based on the predicted similarity of each frame. For example, SipMask \cite{Cao_SipMask_ECCV_2020} generates spatial coefficients for each instance on its detector to deploy temporal context from multiple frames. VisTR \cite{wang2021end} introduces the Transformer to VIS in an offline manner and models the instance queries for the whole video. CrossVIS \cite{yang2021crossover} proposes an instance-to-pixel relation learning scheme, which localizes the same instances in other frames based on the instance features in the current frame. IFC \cite{NEURIPS2021_6f2688a5} significantly relaxes the heavy computation and memory usage via the proposed inter-frame communication transformers. SeqFormer \cite{10.1007/978-3-031-19815-1_32} utilizes a query decomposition mechanism to dynamically allocate spatial attention on each frame and perform query-level temporal aggregation. TeViT \cite{Yang_2022_CVPR} proposes the messenger shift mechanism and spatiotemporal query interaction mechanism to model frame-level and instance-level temporal context information. EfficientVIS \cite{Wu_2022_CVPR} is an online method that proposes correspondence learning to link adjacent instance tracklets to eliminate hand-crafted data association. These multiple-stage approaches require the design and tuning of a separate model to improve overall performance and rely heavily on image-level instance segmentation models and complex human-designed rules to associate instances.

In contrast, single-stage VIS jointly tracks and segments instances in an end-to-end manner. STEm-Seg exploits a video clip as a single 3D spatiotemporal volume, enhancing the feature representation of spatiotemporal and improving performance through instance position and then separating object instances by clustering learned embeddings \cite{liang2024clusterfomer, wu2019sequence, cui2024collaborative, venugopalan2015sequence}. However, the quality of masks generated based solely on position information is limited. To address this limitation, a novel mixing distribution is proposed to promote the instance boundary by fusing hierarchical appearance embedding in a coarse-to-fine manner.

\noindent\textbf{Online video analysis.} Besides focusing on the performance, recent works \cite{jiang2020learning, Xu2020CenterNetHP, cui2022dfa} also begin to improve the efficiency of the video object detectors for online applications. For example, CenterNet-HP \cite{Xu2020CenterNetHP} extends CenterNet \cite{duan2019centernet} for the video domain tasks to replace the two-stage detectors like Faster R-CNN \cite{ren2015faster}. LSTS \cite{jiang2020learning} proposes a learnable spatial-temporal sampling strategy to efficiently and effectively aggregate the features from the neighboring frames based on high-level semantics. DFA \cite{cui2022dfa, cui2022dynamic} introduces two modules that dynamically choose the number of frames used for feature aggregation. For online video analysis \cite{guo2022ec2detect, fujitake2022video}, most of the existing post-processing-based models \cite{zhu2017deep, ren2015faster, deng2019object, feichtenhofer2017detect} can be adapted to the online version. Besides these models, Lu et al. \cite{Lu_2017_ICCV} introduce an association LSTM module to refine the predictions with the consecutive frames. DorT \cite{luo2019detect} proposes a strategy to determine the video frames with object tracking or detection. GMLCN \cite{han2022global} applies a global memory bank to enhance the object features and designs an object tracker to reduce redundant detection processes to balance the effectiveness and efficiency during the inference process. 

\noindent\textbf{Global-local aggregation.} To improve the local granularity and global semantics, global-local aggregation is widely used in the computer vision tasks \cite{liu2021densernet, sarlin2019coarse,sun2021loftr, CUI2021300, ding2019camnet, lin2021gait, dong2022motion, yan2022gl, yan2021hierarchical}. For the video object detection task, MEGA \cite{chen2020memory} introduces a strategy to aggregate the temporal information with similar semantics and close temporal relations so that both global and local features are fused to enhance the object representations. TF-Blender \cite{cui2021tf} distinguishes the spatial global and local features and fuses them by generating a temporal-feature graph. Recently, Transformers-based models, like TransVOD \cite{zhou2022transvod, he2021end} and PTSEFormer \cite{wang2022ptseformer},  introduce multi-head attention modules to video object detection to aggregate the feature representations globally and locally in both temporal and spatial domains. However, these models equally treat the global and local features from different frames. In our models, the current frame to be aggregated provides the global information, while the neighboring frames introduce the local features, which are different from the existing ones.

\noindent\textbf{Multi-stage models in object detection.} Multi-stage models are widely used techniques to improve object detection performance. R-CNN families \cite{gao2021fast, ren2015faster, he2017mask} first set up the pipeline for two-stage object detectors where proposals are roughly predicted and then refined to generate the final outputs. Following these works, Cascade R-CNN \cite{cai2019cascade} introduces a multi-stage pipeline to solve the problems of overfitting during training and a quality mismatch at inference time. Recently, query-based object detectors \cite{sun2021sparse, gao2022adamixer, detr, zhu2021deformable, gao2021fast, liu2022dab, roh2021sparse, li2022dn, zhang2022dino}, especially Transformer-based models, treat the object detection problem as a new perspective. DETR \cite{detr} first introduces a sequence-to-sequence model with multi-stage of Transformer encoder-decoder layers to generate the final predictions according to $100$ randomly initialized queries. Following DETR, REGO-Deformable-DETR\cite{chen2022recurrent} is the most similar to our works, which gradually refines the prediction results in the existing Transformer-based models. Unlike this work, our models are designed for the video object detection task. We introduce a dynamic version to adaptively and reconfigurably refine the predictions according to the neighboring frames. 

\noindent\textbf{\textcolor{black}{Autonomous driving}}\textcolor{black}{, a transformative paradigm in the realm of transportation, represents a convergence of cutting-edge technologies that aim to redefine the future of mobility \cite{liu15tnnls, liu2018mobile, cui2022dg}. At the core of this paradigm shift is the vision to create vehicles capable of navigating and interacting with their environments without human intervention. This ambitious goal has spurred extensive research across various disciplines, ranging from computer science and artificial intelligence to robotics and automotive engineering \cite{liu2020indoor, liu2020video, liu15tnnls}. The pursuit of autonomous driving is motivated by a vision of safer, more efficient, and environmentally friendly transportation systems. By leveraging advanced sensor technologies, machine learning algorithms, and real-time data processing, researchers and engineers strive to create intelligent vehicles capable of perceiving and interpreting their surroundings with a level of sophistication comparable to, if not exceeding, human capabilities \cite{lin2004automatic, liu2022prophet, wang2022towards}.
}

\textcolor{black}{Video object detection plays a crucial role in the development and implementation of autonomous driving systems \cite{liu2020video, wu2019sequence, chen2020memory}. Autonomous vehicles rely on various sensors, including cameras, to perceive their surroundings and make informed decisions. Video object detection specifically involves identifying and tracking objects within a video stream, which is essential for understanding the dynamic environment around the vehicle \cite{redmon2018yolov3, liu2016ssd}. Here are some key points that highlight the relationship between video object detection and autonomous driving: }

\textcolor{black}{1. Perception and Awareness: Video object detection enables the vehicle to perceive and understand the environment by identifying and tracking objects such as pedestrians, vehicles, cyclists, and obstacles in real time \cite{cui2022dfa, tan2019efficientnet, sun2022efficient}. Accurate and timely detection is critical for the vehicle to have a comprehensive awareness of its surroundings, allowing it to make informed decisions.
2. Sensor Fusion:  Autonomous vehicles typically integrate data from multiple sensors, including cameras, LiDAR (Light Detection and Ranging), radar, and others.
Video object detection complements other sensor data, providing rich visual information that helps in building a more comprehensive and robust perception system through sensor fusion \cite{yan2021hierarchical, henschel2018fusion, lin2020dual, WangControllable}. 3. Decision-Making: The information obtained from video object detection contributes to the decision-making process of the autonomous vehicle. For instance, recognizing a pedestrian crossing the road might trigger the vehicle to slow down or stop. Understanding the context of the detected objects is crucial for making safe and efficient driving decisions \cite{liu2020large, jiang2019video}. 4. Real-Time Adaptation: Autonomous vehicles need to adapt to dynamic and changing environments. Video object detection algorithms must operate in real-time to continuously analyze the incoming video feed and update the vehicle's understanding of the surroundings. Real-time adaptation is essential for handling scenarios such as sudden obstacles or unexpected events \cite{guo2022virtual, wojke2017simple}.}

\textcolor{black}{In summary, video object detection is a foundational element in the perception stack of autonomous vehicles, providing critical information for understanding and navigating the complex and dynamic driving environment. Advances in this technology are integral to the ongoing progress and deployment of autonomous driving systems.}
\section{Methodology}
Our proposed video object detectors follow a step-by-step refinement pipeline. In this section, we first review the current Transformer-based video object detectors. Then, we introduce our stepwise spatial global-local aggregation networks to gradually refine the prediction results according to the neighboring frames. Our proposed methods include a multi-stage stepwise network, spatial global-local aggregation, and dynamic aggregation strategy. We will elaborate on them in detail in the following sections.
\subsection{Preliminary}
Given the input frame $\bm{I}$, its corresponding features extracted from the backbone like ResNet \cite{he2016deep} are represented as $\bm{F}$, which are then fed into an intermediate network $\mathcal{H}$ to enhance the features used for object detection. For two-stage object detectors, $\mathcal{H}$ represents the region proposal networks \cite{ren2015faster} together with the region of interests operations \cite{ren2015faster, he2017mask}. For Transformer-based object detectors, $\mathcal{H}$ denotes the sequence-to-sequence Transformer encoder and decoder layers. The features after the network $\mathcal{H}$ are then fed into the task head $\mathcal{N}_{cls}$ and $\mathcal{N}_{box}$ for the object category classification and position regression, respectively. The process of the existing object detectors can be summarized as:

\begin{equation}
    \begin{aligned}
    \bm{C} &= \mathcal{N}_{cls}\left(\mathcal{H}\left(\bm{F}\right)\right)\\
    \bm{B} &= \mathcal{N}_{box}\left(\mathcal{H}\left(\bm{F}\right)\right),
    \end{aligned}
    \label{eq: detr}
\end{equation}
where $\bm{C} \in \mathbb{R}^{n\times c}, \bm{B} \in \mathbb{R}^{n\times 4}$ represent the category and location of the bounding boxes and $n, c$ denote the number of objects and categories respectively. 

We use $\bm{I}_i$, with the corresponding feature $\bm{F}_i$, to represent the neighboring frames of $\bm{I}$, where $\forall\bm{I}_i \in \mathcal{N}\left(\bm{I}\right)$ and the size of $\mathcal{N}\left(\bm{I}\right)$ is $l$. 
For feature-aggregation based models, they fuse the features $\bm{F}$ from the neighboring frames to get the aggregated outputs $\Delta\bm{F}$, as:

\begin{equation}
    \Delta\bm{F} = \sum_{\forall\bm{I}_i\in\mathcal{N}\left(\bm{I}\right)}w_{i}\bm{F}_i,
\end{equation}
where $w_{i}$ denotes the weights used for aggregation. These methods mentioned above follow the same pattern in that all the neighboring frames are taken into account at once during the post-processing or feature-aggregation processes. Unlike these models, we propose a module to aggregate and refine the predictions step by step. 

 \begin{figure}[!bt]
    \centering
    \includegraphics[width=7cm]{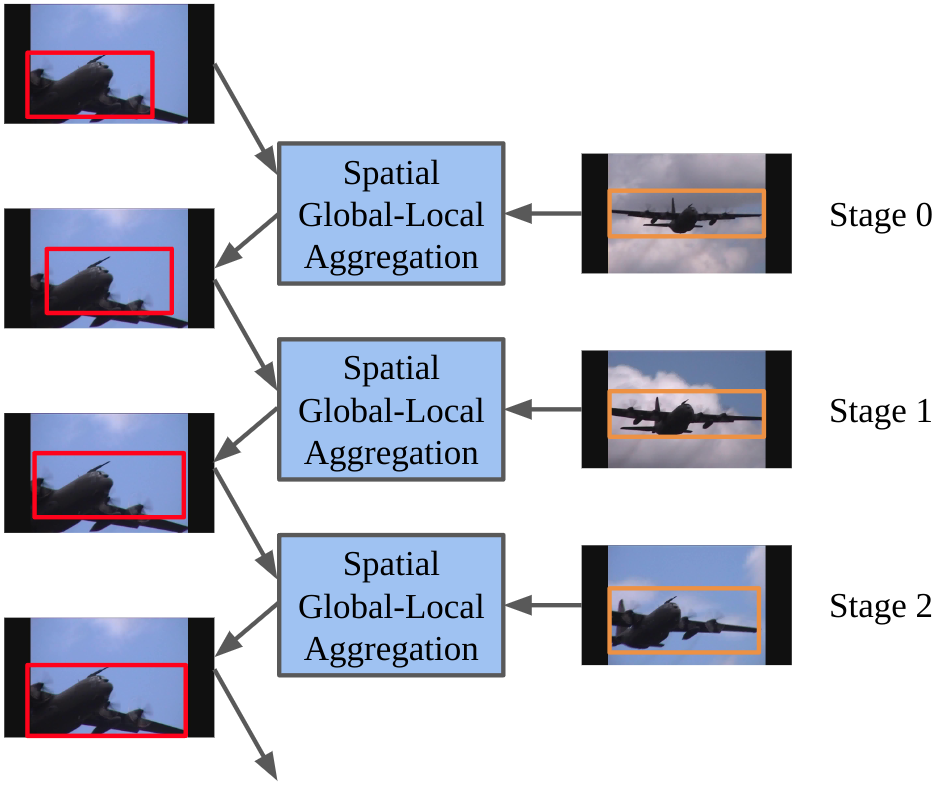}
    \caption{Framework of the multi-stage stepwise network. The bounding boxes in the current frame (marked as red boxes) are gradually refined according to the prediction results from the neighboring frames (marked as orange boxes) using the spatial global-local aggregation module.}
    \label{fig: framework}
\end{figure}

\subsection{Multi-stage stepwise Network}
We notice several issues caused by the aggregation process in the current feature-aggregation-based models \cite{zhu2017flow, chen2020memory, wang2018fully, cui2021tf, deng2019relation, wu2019sequence, gong2021temporal}, which make them unsuitable for the online video applications. To start with, They usually choose a large number of neighboring frames for aggregation. Though there are recent works like DFA \cite{cui2022dfa}, which dynamically select the number of frames, they still aggregate all the features at one time, which significantly increases the model's computational complexity. Moreover, the performance of those methods will drop a lot when the number of frames used for aggregation is small. For example, for FGFA \cite{zhu2017flow} with ResNet-50 as the backbone, the \textcolor{black}{$\text{AP}_{50}$} will drop from $74.1\%$ to $72.3\%$ when the number of frames decreases from $25$ to $5$; for TransVOD++ \cite{zhou2022transvod} with Swin Base \cite{liu2021swin} as the backbone, the \textcolor{black}{$\text{AP}_{50}$} will be improved from $88.0\%$ to $90.0\%$ when the number of frames increases from $2$ to $14$. Furthermore, those methods rely on the information from future frames to keep the current performance. Take FGFA \cite{zhu2017flow} as an example; the performance of \textcolor{black}{$\text{AP}_{50}$} will drop from $74.1\%$ to $72.9\%$ when only the current and previous frames are used for aggregation.

\textcolor{black}{
\begin{table*}[!bt]
    \centering
    \begin{tabular}{c|c}
    \toprule
    Notation & Meaning  \\
    \midrule
    $\bm{I}$ &  Input frame \\
    $\bm{F}$ & Features extracted from the backbone for input frame $\bm{I}$\\
    $\mathcal{H}$ & Intermediate network to enhance the features used for object detection\\
    $\mathcal{N}_{cls}$ & Network for object category classification\\
    $\mathcal{N}_{box}$ & Network for object position regression \\
    $\bm{C}$ & Category of the bounding boxes\\
    $\bm{B}$ & Location of the bounding boxes\\
    $c$ & Number of object categories \\
    $n$ & Number of objects to be detected\\
    $\bm{I}_i$ & Neighboring frame \\
    $\bm{F}_i$ &  Features extracted from the backbone for neighboring frame $\bm{I}_i$\\
    $l$ & Number of neighboring frames \\
    $\mathcal{N}(\bm{I})$ & Neighboring frame set\\
    $\Delta \bm{F}$ & Aggregated feature from the neighboring frames in feature aggregation based models\\
    $w_i$ & Weights used for aggregating features from the neighboring frames \\
    $\bm{C}_i$ &  Category of the bounding boxes of $i$-th stage / frame\\
    $\bm{B}_i$ & Location of the bounding boxes of $i$-th stage / frame\\
    $\alpha$ & Hyperparameters to adjust $\bm{B}_i$\\
    $\Delta \bm{E}_i$ & Aggregated feature from the neighboring frames in our proposed model at stage $i$\\
    $\mathcal{S}$ & Sampling function\\
    $\mathcal{A}$ & Aggregation module to generate $\Delta \bm{E}_i$ \\
    $\bm{G}_i$ & Globally aggregated feature at $i$-th stage\\
    $\bm{L}_i$ & Locally aggregated feature at $i$-th stage\\
    $\mathcal{P}$ & Mapping function \\
    $\mathcal{R}$ & ROI operation\\ 
    $\beta_i$ & The scale factor for $\mathcal{R}$\\
    $\mathcal{M}$ & The multi-head cross attention \\
    $\sigma$ & Mapping function\\
    $\delta$ & Threshold for dynamic aggregation \\
     \bottomrule
    \end{tabular}
    \caption{Notations used in the work}
    \label{tab: notation}
\end{table*}}

We introduce a multi-stage stepwise network as Figure \ref{fig: framework} to handle the issues mentioned above. \textcolor{black}{Meanwhile, we list all the notations used in the work in Table \ref{tab: notation}.} Inspired by Cascade R-CNN \cite{cai2019cascade} and Sparse R-CNN \cite{sun2021sparse}, which refines the object detection predictions step by step, we introduce a multi-stage processing pipeline to gradually improve the performance of video object detection with the help of the neighboring frames. In detail, given $l$ neighboring frames used for aggregation, we apply $l$ stages of stepwise aggregation networks to improve the predictions step by step. Each stage takes the prediction results and object representations from one of the neighboring frames to refine the outputs and features of the current generated bounding boxes. The refined results will serve as the inputs for the following stage. The process can be summarized as follows:


\textcolor{black}{
\begin{equation}
    \begin{aligned}
    \bm{C}_{i}  &= \mathcal{N}_{cls}\left(\Delta\bm{E}_i \right)\\
    \bm{B}_i  &= \mathcal{N}_{box}\left(\Delta\bm{E}_i \right) + \alpha \bm{B}_{i - 1} ,
    \end{aligned}
\end{equation}
}
\textcolor{black}{
where $\bm{C}_{i}, \bm{B}_{i} $ denote the prediction results of object categories and locations at the $i$-th stage using the $i$-th neighboring frame and $\alpha$ is a hyperparameter, and $\Delta\bm{E}_i $ represents the refined features at the current $i$-th stage, where $\forall\bm{I}_i \in \mathcal{N}\left(\bm{I}\right)$. In terms of the order of $\bm{I}_i$ and $\bm{F}_i$, we have a sampling function $\mathcal{S}$ to determine the sequence and will be discussed in the experiment sections. By default, $\mathcal{S}$ is set to be a sorting function with descending orders, where frames with similar semantics or close to the current frame are the last used for the stepwise aggregation. To achieve $\Delta\bm{E}_i $, we use the following equation:}

\textcolor{black}{
\begin{equation}
    \Delta\bm{E}_i  = \mathcal{A}\left(\Delta\bm{E}_{i-1} ,\bm{F}_i, \bm{B}_i \right),
\end{equation}
}
\textcolor{black}{
where $\mathcal{A}$ is the aggregation module, which will be discussed in detail in the next section. For the first stage where $i=0$, we use the predictions and features of the current frame without aggregation, represented as $\bm{C}_0 =\bm{C}, \bm{B}_0 =\bm{B}$ and $\Delta\bm{E}_0 =\mathcal{H}\left(\bm{F}\right)$ in detail, where $\mathcal{H}$ maps the feature map $\bm{F}$ into a vector.} 

 \begin{figure*}[!bt]
    \centering
    \includegraphics[width=15cm]{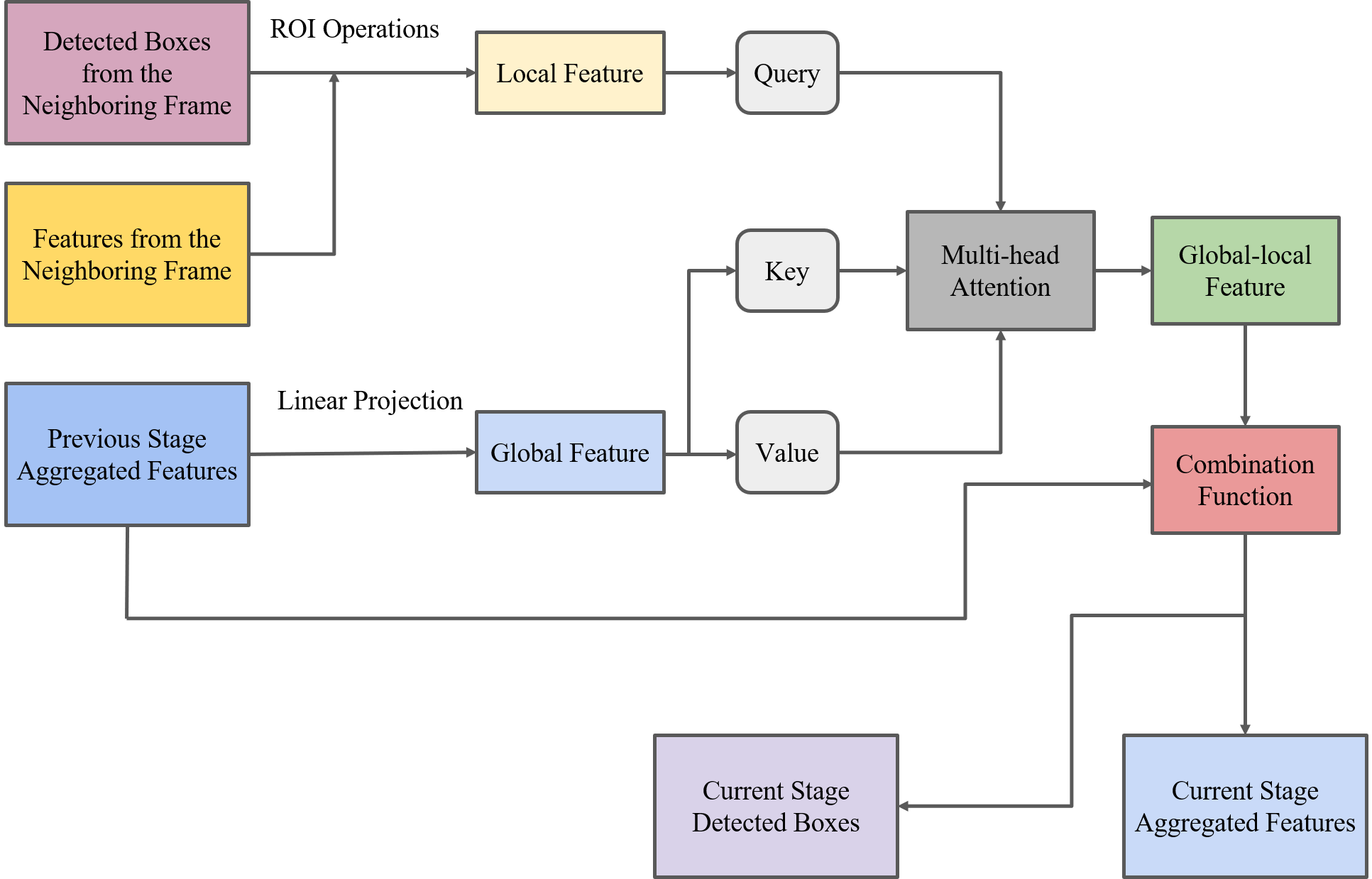}
    \caption{Framework of spatial global-local aggregation.}
    \label{fig: framework2}
\end{figure*}

\subsection{Spatial Global-local Aggregation}
When extending the image object detectors to the video domain tasks, researchers pay more attention to the temporal global-local information to improve the performance. TF-Blender \cite{cui2021tf} is one of the few works which takes spatial information into account during the feature aggregation process. Transformer-based video object detectors \cite{wang2022ptseformer, he2021end, zhou2022transvod} usually combine the spatial and temporal features globally and locally. However, their complicated and well-designed aggregation modules are neither time-efficient nor reconfigurable, which makes it difficult for online applications. 

To overcome the problems, we introduce a spatial global-local aggregation module. The idea is that when the current frame $\bm{I}$ is with feature degradation, we want to combine the global features from $\bm{I}$ and the local ones from $\bm{I}_i$ to improve the object representations. Given a clip of videos, most of the frames share similar high-level semantics since the numbers and categories of objects within a video do not change abruptly. Different from the global-level semantics, local information varies significantly at different frames. In our spatial global-local aggregation module, we combine the accurate, fine-grained local features from the neighboring frames and coarse global information from the current frame with feature degradation to improve the object representations for the final outputs. 


\textcolor{black}{In detail, at the $i$-th stage, we refine the prediction results $\bm{B}_{i-1} $ and feature representation $\Delta{E}_i $ from the previous stage with the guide of the neighboring frame outputs $\bm{B}_i$ and their corresponding features $\Delta\bm{E}_{i-1} $ and $\bm{F}_i$. Global features, denoted as $\bm{G}_{i}$, are mapped the from refined representations $\Delta\bm{E}_{i-1} $ in the previous stage and local feature $\bm{L}_{i}$ are provided by the current neighboring frame $\bm{B}_i$ and $\bm{F}_i$. For implementation, we use a mapping function $\mathcal{P}$ to project $\Delta\bm{E}_{i-1} $ to generate the global feature $\bm{G}_i$. For the local feature $\bm{L}_i$, we apply region of interest-based operations like ROI Align \cite{he2017mask} to extract the representation as:}

\textcolor{black}{
\begin{equation}
\begin{aligned}
    \bm{G}_i & = \mathcal{P}\left(\Delta\bm{E}_{i-1} \right)\\
    \bm{L}_i &= \mathcal{R}\left(\bm{B}_i, \bm{F}_i, \beta_i\right), 
    \label{eq: align}
\end{aligned}
\end{equation}
}
where $\mathcal{R}$ denotes the ROI operation and $\beta_i$ represents the scale factor used for $\mathcal{R}$. 
\begin{table*}[!bt]
    \centering
    \begin{tabular}{l|l|l|l|l|l|l|l}
    \toprule
    Method & mAP & AP$_\text{50}$ & AP$_\text{75}$ & AP$_\text{S}$ & AP$_\text{M}$ & AP$_\text{L}$ & FPS \\
    \midrule
    \midrule
    \multicolumn{7}{l}{Two-stage Object Detectors}\\
    \midrule
    Faster R-CNN\cite{ren2015faster} + DFF \cite{zhu2017deep}  & 42.7 & 70.3 & 45.7 & 5.0 & 17.6 & 48.7 & 17.3 \\
    Faster R-CNN\cite{ren2015faster} + FGFA \cite{zhu2017flow} & 47.1 & 74.7 & 52.0 & 5.9 & 22.2 & 53.1 & 14.9\\
    Faster R-CNN\cite{ren2015faster} + SELSA \cite{wu2019sequence} & 48.7 & 78.4 & 53.1 & 8.5 & 26.3 & 54.5 & 14.1\\
    Faster R-CNN\cite{ren2015faster} + Temporal RoI Align \cite{gong2021temporal} & 48.5 & 79.8 & 52.3 & 7.2 & 26.5 & 54.4 & 10.5\\
    \textcolor{black}{Faster R-CNN \cite{ren2015faster} + Ours ($\delta=0.9$)} & 49.1 & 81.2 & 52.8 & 7.9 & 27.3 & 54.9 & 16.1 \\
    \textcolor{black}{Faster R-CNN \cite{ren2015faster} + Ours ($\delta=0.8$)} & 48.7 & 80.3 & 52.1 & 7.4 & 26.8 & 54.6 & 17.0\\
    \midrule
    \midrule
    \multicolumn{7}{l}{DETR-based Object Detectors}\\
    \midrule
    Conditional-DETR\cite{Meng_2021_ICCV} & 53.7 & 74.7 & 60.1 & 7.7 & 25.9 & 60.6 & 13.1\\
    Conditional-DETR\cite{Meng_2021_ICCV} + Ours & 55.1$_{\uparrow1.4}$ & 76.0$_{\uparrow1.3}$ & 62.0$_{\uparrow1.9}$ & 9.8$_{\uparrow2.1}$ & 28.0$_{\uparrow2.1}$ & 62.0$_{\uparrow1.4}$ & 12.3\\
    \midrule
    SMCA-DETR\cite{gao2021fast} & 53.5 & 74.2 & 59.6 & 7.6 & 25.7 & 60.5 & 13.4\\
    SMCA-DETR\cite{gao2021fast} + Ours & 54.7$_{\uparrow1.2}$ & 75.7$_{\uparrow1.5}$ & 61.8$_{\uparrow2.2}$ & 9.5$_{\uparrow1.9}$ & 27.5$_{\uparrow1.8}$ & 61.6$_{\uparrow1.1}$ & 12.7\\
    \midrule
    DAB-DETR\cite{liu2022dab} & 54.2 & 75.3 & 61.3 & 8.9 & 26.8 & 61.2 & 12.0\\
    DAB-DETR\cite{liu2022dab} + Ours & 55.6$_{\uparrow1.4}$ & 76.5$_{\uparrow1.2}$ & 63.2$_{\uparrow1.9}$ & 10.2$_{\uparrow2.3}$ & 29.3$_{\uparrow2.5}$ & 62.8$_{\uparrow1.6}$ & 11.5\\
    \midrule
    Deformable-DETR\cite{zhu2021deformable} & 55.4 & 76.2 & 62.2 & 10.5 & 27.5 & 62.3 & 15.3\\
    Deformable-DETR\cite{zhu2021deformable} + TransVOD\cite{he2021end} & 56.1$_{\uparrow0.7}$ & 76.6$_{\uparrow0.4}$ & 63.3$_{\uparrow1.1}$ & 11.6$_{\uparrow1.1}$ & 28.8$_{\uparrow1.3}$ & 63.1$_{\uparrow0.8}$ & 13.9 \\
    Deformable-DETR\cite{zhu2021deformable} + TransVOD++\cite{zhou2022transvod} & 56.2$_{\uparrow0.8}$ & 76.8$_{\uparrow0.6}$ & 63.7$_{\uparrow1.5}$ & 12.2$_{\uparrow2.7}$ & 29.5$_{\uparrow2.0}$ & 63.2$_{\uparrow0.9}$ & 13.3\\
    Deformable-DETR\cite{zhu2021deformable} + TransVODLite\cite{zhou2022transvod} & 55.8$_{\uparrow0.4}$ & 76.5$_{\uparrow0.3}$ & 62.9$_{\uparrow0.7}$ & 11.2$_{\uparrow0.7}$ & 28.1$_{\uparrow0.6}$ & 62.9$_{\uparrow0.6}$ & 14.3\\
    \textcolor{black}{Deformable-DETR\cite{zhu2021deformable} + Ours ($\delta=0.9$)} &  57.0$_{\uparrow1.6}$ & 77.2$_{\uparrow1.0}$ & 64.2$_{\uparrow2.0}$ & 13.0$_{\uparrow2.5}$ & 30.1$_{\uparrow2.6}$ & 63.6$_{\uparrow1.3}$ & 14.1\\
    \textcolor{black}{Deformable-DETR\cite{zhu2021deformable} + Ours ($\delta=0.8$)}&  56.8$_{\uparrow1.4}$ & 76.9$_{\uparrow0.7}$ & 64.0$_{\uparrow1.8}$ & 12.5$_{\uparrow2.0}$ & 29.8$_{\uparrow2.3}$ & 63.3$_{\uparrow1.0}$ & 14.9\\
    \midrule
    \bottomrule
    \end{tabular}
    \caption{Performance comparison with the recent state-of-the-art video object detection approaches on ImageNet VID validation set \cite{russakovsky2015imagenet}. The backbone is ResNet-50 \cite{he2016deep}.}
    \label{tab: mainResult}
\end{table*}

\begin{table*}[!bt]
    \centering
    \begin{tabular}{l|l|l|l|l|l|l|l}
    \toprule
    Method & mAP & AP$_\text{50}$ & AP$_\text{75}$ & AP$_\text{S}$ & AP$_\text{M}$ & AP$_\text{L}$ & FPS \\
    \midrule
    \textcolor{black}{Deformable-DETR\cite{zhu2021deformable}} & \textcolor{black}{66.7} & \textcolor{black}{90.3} & \textcolor{black}{75.4} & \textcolor{black}{22.3} & \textcolor{black}{40.7} & \textcolor{black}{75.4} & \textcolor{black}{11.5}\\
    \textcolor{black}{Deformable-DETR\cite{zhu2021deformable} + TransVOD\cite{he2021end}} & \textcolor{black}{67.2$_{\uparrow0.5}$} & \textcolor{black}{90.8$_{\uparrow0.5}$} & \textcolor{black}{76.4$_{\uparrow1.0}$} & \textcolor{black}{23.1$_{\uparrow0.8}$} & \textcolor{black}{41.4$_{\uparrow0.7}$} & \textcolor{black}{75.8$_{\uparrow0.4}$} & \textcolor{black}{10.7} \\
    \textcolor{black}{Deformable-DETR\cite{zhu2021deformable} + TransVOD++\cite{zhou2022transvod}} & \textcolor{black}{67.5$_{\uparrow0.8}$} & \textcolor{black}{90.8$_{\uparrow0.5}$} & \textcolor{black}{76.6$_{\uparrow1.2}$} & \textcolor{black}{23.3$_{\uparrow1.0}$} & \textcolor{black}{41.6$_{\uparrow0.9}$} & \textcolor{black}{76.1$_{\uparrow0.7}$} & \textcolor{black}{10.1}\\
    \textcolor{black}{Deformable-DETR\cite{zhu2021deformable} + TransVODLite\cite{zhou2022transvod}} & \textcolor{black}{67.6$_{\uparrow0.9}$} & \textcolor{black}{91.0$_{\uparrow0.7}$} & \textcolor{black}{76.8$_{\uparrow1.4}$} & \textcolor{black}{23.4$_{\uparrow1.1}$} & \textcolor{black}{41.6$_{\uparrow0.9}$} & \textcolor{black}{76.2$_{\uparrow0.8}$} & \textcolor{black}{11.3}\\
    \textcolor{black}{Deformable-DETR\cite{zhu2021deformable} + Ours} &  \textcolor{black}{68.1$_{\uparrow1.4}$} & \textcolor{black}{90.4$_{\uparrow1.1}$} & \textcolor{black}{76.9$_{\uparrow1.5}$} & \textcolor{black}{23.8$_{\uparrow1.5}$} & \textcolor{black}{42.1$_{\uparrow1.4}$} & \textcolor{black}{76.5$_{\uparrow1.1}$} & \textcolor{black}{11.2}\\
    \midrule
    \bottomrule
    \end{tabular}
    \caption{Performance comparison with the recent state-of-the-art video object detection approaches on DGL-MOTS validation set \cite{russakovsky2015imagenet}. The backbone is ResNet-50 \cite{he2016deep}.}
    \label{tab: mainResultKITTI}
\end{table*}

Inspired by REGO-Deformable-DETR \cite{chen2022recurrent}, we apply multi-head cross-attention \cite{vaswani2017attention} to model the relations between the global and local features. Following the skip connection operations in ResNet \cite{he2016deep}, we apply a combination function $\mathcal{G}$ to fuse aggregated global-local features with the aggregated features from the previous stage, as follows:


\textcolor{black}{
\begin{equation}
\begin{aligned}
    \Delta\bm{E}_i  &= \mathcal{A}\left(\Delta\bm{E}_{i-1} ,\bm{F}_i, \bm{B}_i \right) \\
    &= \sigma\left(\mathcal{G}\left(\mathcal{M}\left(\bm{L}_i, \bm{G}_{i} \right), \Delta\bm{E}_{i - 1} \right)\right) \\
    &= \sigma\left(\mathcal{G}\left(\mathcal{M}\left(\mathcal{R}\left(\bm{B}_i, \bm{F}_i, \beta_i\right),  \mathcal{P}\left(\Delta\bm{E}_{i-1} \right) \right)\right),  
    \Delta\bm{E}_{i - 1}\right) \\
\end{aligned}
\end{equation}
}
\textcolor{black}{where $\mathcal{M}$ represents multi-head cross attention \cite{vaswani2017attention} and $\bm{L}_i$ serves as the query vector and $\bm{G}_{i} $ is used for the key and value vectors. $\sigma$ is a mapping function.}
\subsection{Dynamic Aggregation Strategy}
In the stepwise aggregation network mentioned above, the number of stages is the same as that of the frames used for aggregation and always fixed regardless of the input frames. However, there is no need to always use a large number of stages for the frames without feature degradation, like the objects with slow motion speeds mentioned in FGFA \cite{zhu2017flow} and DFA \cite{cui2022dfa}. Inspired by DFA \cite{cui2022dfa}, we introduce a dynamic aggregation strategy where the number of stages is adaptive to the input frames.

Unlike the strategies used in DFA \cite{cui2022dfa}, which determines the model computational complexity according to the estimated object sizes and motion speeds, our method is based on the outputs of two consecutive stages. For stage $i$ and its previous stage $i - 1$, if the output aggregated features $\Delta\bm{E}_i$ and $\Delta\bm{E}_{i - 1}$ are similar enough, the model will not go to the next stage $i + 1$. During the training process, we take all the stages into account. For the inference, when the cosine similarity between two consecutive stages is more than the criteria threshold $\delta$ which is set to $0.9$ by default, the model will stop the aggregation process.

Meanwhile, the dynamic aggregation strategy solves a problem that most of the existing feature-aggregation-based models encounter: lack of reconfigurability. For models like TransVOD \cite{zhou2022transvod, he2021end}, and PTSEFormer \cite{wang2022ptseformer}, a new model needs to be retrained when the number of frames used for aggregation is changed during the inference time. On the contrary, our method does not require a newly trained model but still works for a different number of frames by simply changing $\delta$ from $0.9$ to other values.

\subsection{Online Application Properties}
Here, we summarize the properties of our proposed models for online video applications.

\noindent\textbf{Effectiveness.} Our proposed model can improve the performance of the original image object detectors on the video domain tasks by gradually refining the predictions and representations of each stage.

\noindent\textbf{Efficiency.} Our model only relies on the prediction results and their corresponding features from the previous frames. Therefore, when implemented with a queue-like memory bank, our model does not need to calculate the information provided by the previous frames multiple times, which is more efficient than most of the current Transformer-based video object detectors like TransVOD \cite{zhou2022transvod, he2021end} and PTSEFormer \cite{wang2022ptseformer}.

\noindent\textbf{Reconfigurability.} As mentioned in the dynamic aggregation strategy section, when changing the number of frames used for aggregation, our model can be easily reconfigured by reloading different criteria values to stop the aggregation process without retraining a new model. 

\section{Experiments}
\subsection{Experiment Setup}
\noindent\textbf{Experiment details.} We evaluate our proposed methods on the ImageNet VID benchmark \cite{russakovsky2015imagenet} with the recent state-of-the-art Transformer-based object detection models \cite{gao2021fast, zhu2021deformable, liu2022dab, Meng_2021_ICCV}. Following the pipeline of TransVOD \cite{zhou2022transvod, he2021end}, we first pretrain our models on the MS COCO \cite{lin2014microsoft} and then fine-tune on the combination of ImageNet VID and DET datasets. \textcolor{black}{Besides the ImageNet VID benchmark, we also conducted experiments on car-related autonomous driving datasets like DGL-MOTS \cite{cui2022dg, 10.1145/3632181}. } All the models are trained on $8$ Tesla A100 GPUs, and during the training and inference processes, $4$ neighboring frames are used for aggregation. 

\noindent\textbf{Implementation details.} The number of neighboring frames/stages $n$ is set to $4$. All the mapping functions like  $\mathcal{P}$ and $\sigma$ are implemented as MLPs. $\mathcal{R}$ is implemented with ROI Align \cite{he2017mask}. For the other details, we follow the exact implementation as REGO-Deformable-DETR \cite{chen2022recurrent}.  Following the pipeline of Deformable-DETR \cite{zhu2021deformable}, Conditional-DETR \cite{Meng_2021_ICCV}, SMCA-DETR \cite{gao2021fast} and DAB-DETR \cite{liu2022dab}, ImageNet\cite{deng2009imagenet, krizhevsky2012imagenet} pre-trained backbones are utilized as the backbones. The number of initial object queries is set as $300$. By default, models are trained for 50 epochs on the MS COCO \cite{lin2014microsoft}, and the learning rate is decayed at the $40$-th epoch by a $0.1$. Following those models mentioned above, we train our models using Adam optimizer \cite{kingma2014adam} with a base learning rate of $2 \times 10^{-4}$, $\beta_1 = 0.9$, $\beta_2 = 0.999$, and weight decay of $10^{-4}$. Learning rates of the linear projections, used for predicting object query reference points and sampling offsets, are multiplied by a factor of $0.1$. 

After pretraining the models on the MS COCO benchmark \cite{lin2014microsoft}, we follow the pipeline of TransVOD \cite{zhou2022transvod, he2021end} and PTSEFormer \cite{wang2022ptseformer} to finetune the pre-trained model on the ImageNet VID benchmark \cite{russakovsky2015imagenet}. We freeze the model beside the classification heads to predict the categories. Following previous work \cite{zhou2022transvod, he2021end, wang2022ptseformer}, we use the same data augmentation, including random horizontal flip, random resizing the input images such that the shortest side is at least $600$ while the longest at most $1000$. We train the network for $7$ epochs, and the learning rate drops at the $5$-th and $6$-th epochs. 

\begin{figure*}[!bt]
    \centering
         \includegraphics[width=13cm]{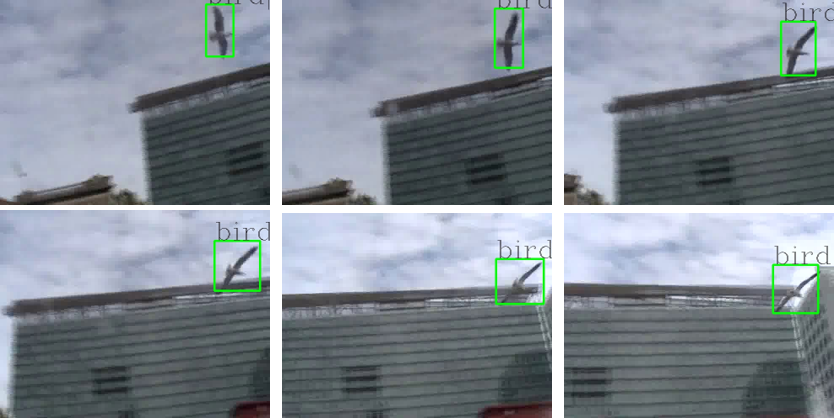}
         \includegraphics[width=13cm]{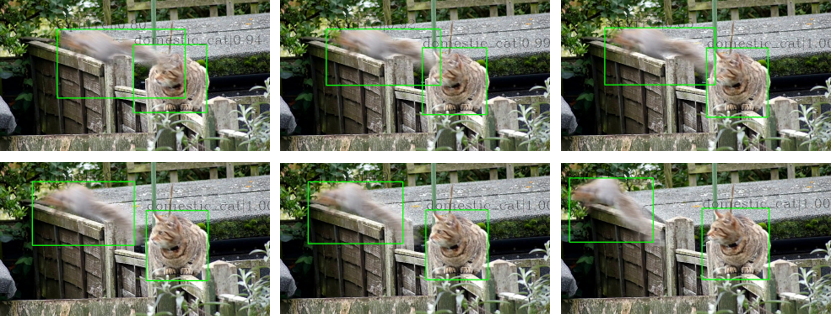}
    \caption{Visualization examples of Deformable-DETR integrated with our proposed methods. The backbone is ResNet-50.}
    \label{fig: visualEx}
\end{figure*}
\begin{table}[!bt]
    \centering
    \begin{tabular}{c|c|c|c|c}
    \toprule
    $\#$ of frames & 1 & 2 & 4 & 8 \\
    \midrule
    mAP & 55.6 & 56.1 & 57.0 & 56.3\\
    \bottomrule
    \end{tabular}
    \caption{Analysis of the number of frames used for aggregation.}
    \label{tab: stageNum}
\end{table}
\subsection{Main Results}
In this section, we conduct experiments with the current Transformer-based object detectors integrated with our proposed modules on the ImageNet VID benchmark \cite{russakovsky2015imagenet}. We compare with the recent state-of-the-art models, summarized in Table \ref{tab: mainResult}. From the table, we notice that when integrated with our proposed modules, the performance on mAP can be increased by at least $1.0\%$. Meanwhile, we find that AP$_\text{75}$ is improved more than AP$_\text{50}$. We argue that our multi-stage stepwise networks can precisely refine the predictions based on the neighboring frames. 

In terms of the object sizes, the objects with smaller sizes perform better than those with larger sizes when integrated with our modules. This is because the global and local features of objects with large sizes are similar, which provides little help to refine the predictions, especially when the objects occupy most of the pixels in the video frame. We visualize some examples with different motion speeds as Figure \ref{fig: visualEx}. From the figure, when integrated with our proposed modules, the model can still accurately detect even objects with a fast motion speed, like birds and squirrels. \textcolor{black}{From the Table \ref{tab: mainResultKITTI}, we can find that on the car-related autonomous driving datasets, our proposed method can also achieve consistent improvement on both the performance and inference speeds.}
\subsection{Model Analysis}
In this section, we conduct experiments on Deformable-DETR \cite{zhu2021deformable} with ResNet-50 as the backbone of the ImageNet VID benchmark \cite{russakovsky2015imagenet} to study the designs of our proposed modules.

\noindent\textbf{Analysis of the number of frames/stages}. We conduct experiments to study how the number of frames/stages will affect the final performance. We use Deformable-DETR \cite{zhu2021deformable} with ResNet-50 as the backbone for the experiments. We select the number of frames from $1$ to $8$, and summarize the results in Table \ref{tab: stageNum}. We notice that the performance can be increased accordingly when increasing the number of frames from $1$ to $4$. However, the performance decreases when the number of frames is $8$. Therefore, different from the current feature-aggregation-based models, which use a large number of neighboring frames, like $14$ for TransVOD \cite{zhou2022transvod,he2021end} or $21$ for FGFA \cite{zhu2017flow, cui2021tf}, our proposed models can achieve the best performance with only a few neighboring frames. We argue that when the number of stages is too big, redundant information will be introduced, which is harmful to refining the predictions.

\noindent\textbf{Analysis of proposed components.} We also conduct experiments on each proposed component in our model to validate their effectiveness, as Table \ref{tab: ablation}. Model A is the original Deformable-DETR \cite{zhu2021deformable} with ResNet-50 as the backbone. We integrate a one-stage spatial global-local aggregation module (SGA) to get model B. By increasing the number of stages from $1$ to $4$, we extend model B to C. By introducing a dynamic aggregation strategy, we get the complete model D. From the table, we notice that when there is only one frame used for aggregation, as model B, the performance can also be improved slightly. When introducing the multi-stage version, the performance can be improved further. Finally, some redundant feature fusion processes can be stopped early using a dynamic aggregation strategy, and the performance is further boosted.

\begin{table}[!bt]
    \centering
    \begin{tabular}{c|c|c|c|c}
    \toprule
    Model & MSN & SGA & DA & mAP \\
    \midrule
    A & & & & 55.4\\
    B & & $\checkmark$ & & 55.6 \\
    C & $\checkmark$ & $\checkmark$ & & \textcolor{black}{56.8}\\
    D & $\checkmark$ & $\checkmark$ & $\checkmark$ & 57.0\\
    \bottomrule
    \end{tabular}
    \caption{Analysis of the contribution of each proposed component.}
    \label{tab: ablation}
\end{table}

\noindent\textbf{Analysis of sampling function $\mathcal{S}$.} We conduct experiments on the choice of sampling function $\mathcal{S}$ as Table \ref{tab: sampling}. ``Ascending" and ``Descending" represent sorting the neighboring frames in ascending or descending orders based on their temporal relations or semantic information. For example, suppose the current frame is the $5-$th of a video, and we would like to sample $4$ frames from the previous frames. Table \ref{tab: sampling} shows an example of a comparison with different sampling functions. From Table \ref{tab: sampling}, we notice that there is not too much difference in the performance when there is no dynamic aggregation strategy (DA). We argue that this is because, finally, all the frames will be used for aggregation, and the order will not affect the final prediction. However, ascending order performs worst when integrated with a dynamic aggregation strategy. We argue that this is because it is easier for the ascending order frames to satisfy the criteria $\sigma$ in dynamic aggregation since these neighboring frames share more similar features.


\begin{table}[!bt]
    \centering
    \begin{tabular}{c|c|c|c|c}
    \toprule
    $\mathcal{S}$ & Ascending & Descending & Random &FPS\\
    \midrule
    Frame Orders & 4,3,2,1 & 1,2,3,4 & 1,3,2,4 &- \\
    \midrule
    mAP (no DA) & 56.7 & 56.8 & 56.6 & \textcolor{black}{13.9}\\
    mAP (with DA) & 56.2 & 57.0 & 56.7 & \textcolor{black}{14.1}\\
    \bottomrule
    \end{tabular}
    \caption{Analysis of the sampling function $\mathcal{S}$.}
    \label{tab: sampling}
\end{table}

\noindent\textbf{Analysis of criteria threshold $\delta$.} We conduct experiments to analyze the effects of the criteria threshold $\delta$ in Table \ref{tab: queries}. By decreasing $\delta$ from $0.9$ to $0.5$, we notice a noticeable drop in the performance. Since $\delta$ is used to determine whether or not to continue the aggregation process, decreasing $\delta$ will force the aggregation process to stop earlier, even when the refined features from the consecutive stages are not similar enough.  


\noindent\textbf{Analysis of the number of queries.} We conduct experiments to analyze the effects of the number of queries as Table \ref{tab: queries}. By increasing the number of queries, the performance can be improved accordingly. However, when there are more than $500$ queries, the performance begins to be saturated. We argue that this is because there are already enough queries, and some of them are redundant.
\begin{table}[! bt]
    \centering
    \begin{tabular}{c|c|c|c|c|c|c|c}
     \toprule   
     & \multicolumn{4}{c|}{$\#$ of Queries } & \multicolumn{3}{c}{$\delta$} \\
     \cline{2-8}
     & 100 & 300 & 500 & 700 & 0.9 & 0.7 & 0.5 \\
     \midrule
     mAP & 56.3 & 57.0 & 57.2 & 56.8 & 57.0 & 56.2 & 55.6\\
    \bottomrule
    \end{tabular}
    \caption{Analysis of the number of queries and $\delta$.}
    \label{tab: queries}
\end{table}

\noindent\textbf{Reconfigurability analysis} We provide more analysis experiments to analyze the reconfigurability of our proposed method. We use Deformable-DETR \cite{zhu2021deformable} with ResNet-50 \cite{he2016deep} as the backbone to analyze the designs of our proposed modules. 
We conduct experiments without the dynamic aggregation strategy for simplicity to achieve this. We train our model with $1, 2, 3$, and $4$ neighboring frames. For the model trained with $4$ neighboring frames, during the inference time, we load different configures to force the model to stop at $1, 2$, and $3$ stages. We compare the results of one model A trained with $4$ neighboring frames while inference with multiple stages with those of multiple models (B, C, D) trained with the corresponding neighboring frames in Table \ref{tab: reconfig}. From the table, we notice that the performance of model A, which is trained with $4$ neighboring frames, when loaded with different configures to inference with different stages, is quite similar to that of the model trained with the corresponding inference stages, which validates the reconfigurability of our propose method.
 
\begin{table}[! bt]
    \centering
    \begin{tabular}{c|c|c|c}
    \toprule
    Model & Training Stages & Inference Stages & mAP \\
    \midrule
    \multirow{4}{*}{A} &  \multirow{4}{*}{4} & 1 & 55.6\\
    & & 2 & 56.1\\
    & & 3 & 56.6\\
    & & 4 & 57.0 \\
    \midrule
    B & 1 & 1 & 55.6\\
    C & 2 & 2 & 56.1\\
    D & 3 & 3 & 56.7\\
    \bottomrule
    \end{tabular}
    \caption{Analysis of reconfigurability}
    \label{tab: reconfig}
\end{table}

\noindent\textbf{Time complexity analysis} We analyze the time complexity of our proposed modules for efficiency. The extra computational complexity introduced by our proposed module is about $50$ GLOPs with $4$ neighboring frames for aggregation, which is below $29\%$ of the original Deformable-DETR \cite{zhu2021deformable}. Compared with the current models for Transformer-based video object detectors, like TransVOD \cite{zhou2022transvod, he2021end} and PTSEFormer \cite{wang2022ptseformer}, which introduce more than $300$ GLOPs (more than $170\%$) to the original Deformable-DETR, the extra computational complexity of our model is neglectable. We noticed that the existing Transformer-based video object detectors like TransVOD and PTSEFormer could not be applied to online applications directly. These models need to predefine the number of neighboring frames used for aggregation in advance and cannot be reconfigured during the inference time. Moreover, these models usually consider future frames during the aggregation process, which is unsuitable for online applications. Finally, these models always have low inference speeds because of the large number of neighboring frames for aggregation. Therefore, we design a different strategy for online video object detection with Transformers. By stepwisely aggregating the neighboring frames, our proposed methods can achieve good performance with little extra computational complexity introduced.

\section{Conclusion}
This work introduces a stepwise spatial global-local aggregation network for online video object detection with Transformers. Our proposed models include three key components: 1.) Multi-stage stepwise network gradually refines the predictions based on the previous results and the information provided by the neighboring frames. 2.) Spatial global-local aggregation module enhances the feature representations by fusing the refined features from the current frame with global semantics and the local information from the neighboring frames with spatial granularity. 3.) Dynamic aggregation strategy stops the fusion process early to improve efficiency and remove redundant operations. The model is trained end-to-end and can be used for most of the existing Transformer-based approaches for online video object detection. We hope our proposed methods can bring some thoughts to the existing works for online video applications.

\bibliographystyle{IEEEtran}
\bibliography{egbib}

\end{document}